\documentclass{article}

\usepackage{iclr2027_conference,times}

\usepackage{amsmath, amssymb}
\usepackage{booktabs}
\usepackage{multirow}
\usepackage{colortbl}
\usepackage{siunitx}
\usepackage{wrapfig}   
\usepackage{tabularx}  
\usepackage{array}

\usepackage{graphicx}
\usepackage{adjustbox}


\usepackage{subcaption}
\usepackage{float}
\usepackage{flafter}
\usepackage[section]{placeins}
\usepackage{wrapfig}
\usepackage{natbib}
\usepackage{hyperref}
\usepackage{url}
\usepackage{enumitem}
\usepackage{csquotes}
\usepackage{listings}
\usepackage{xspace}

\newcommand{\secref}[1]{Sec.~\ref{#1}}
\newcommand{\figref}[1]{Fig.~\ref{#1}}
\newcommand{\tabref}[1]{Table~\ref{#1}}
\newcommand{\appref}[1]{App.~\ref{#1}}
\newcommand{\attack}{JIL\xspace}
\newcommand{\trail}{TRAIL\xspace}

\title{Jumping the Line: Exploiting Length Predictions in LLM Scheduling}

\author{
Yuyang Dai\\
Florida State University\\
\texttt{y9657422@gmail.com}
\And
Rana Shahout\\
MIT\\
\texttt{ranas@mit.edu}
\And
Mahmood Sharif\\
Tel Aviv University\\
\texttt{mahmoods@tauex.tau.ac.il}
}

\iclrfinalcopy

\begin{document}

\maketitle
\fancyhead{}
\lhead{Preprint}

\begin{abstract}

With the proliferation of large language model (LLM) deployments,
efficient request scheduling has become increasingly important for reducing completion time and improving user experience. Size-based scheduling policies, such as Shortest Job First, reduce average completion time by prioritizing shorter requests, but require knowledge or estimates of request sizes. In LLM inference, output lengths, and therefore request sizes, are unknown in advance. As a result, LLM schedulers rely on predicted output lengths to guide scheduling decisions.

This work proposes \attack{}, an attack that exposes a vulnerability in such prediction-based LLM schedulers. By manipulating the scheduler's length prediction, an adversary can cause its requests to receive higher scheduling priority and reduce their completion time. As a case study, we consider \trail{}, a prediction-based LLM scheduling system that uses a lightweight probe to estimate output length. \attack{} exploits this mechanism by optimizing adversarial suffixes that, when appended to a request, cause the probe to underestimate its output length, thereby increasing the request's scheduling priority.

We evaluate \attack{} on two datasets and four LLMs across varied request profiles and deployment configurations. Our results show that \attack{} successfully manipulates the scheduling signal, reducing predicted output lengths by up to 83.4\%. In end-to-end serving experiments, adversarial requests complete up to \(1.53\times\) faster on average. The reduction in predicted length is substantially larger than the change in actual output length, demonstrating a mismatch between the scheduler's estimate and the request's realized size. Response utility varies across models and tasks, sometimes revealing a trade-off between scheduling advantage and response quality. Finally, we evaluate scheduler-side defenses and show that grouping length predictions into coarse intervals reduces \attack{}'s scheduling advantage and mitigates delays to benign requests.

\end{abstract}

\section{Introduction}

Large language models (LLMs) have enabled a new generation of interactive AI applications, such as conversational assistants and coding agents~\cite{ouyang2022training, nam2024using}. In these applications, request completion time, defined as the time from submitting a request until receiving the complete response, directly affects the user experience. As LLM services handle many concurrent requests, efficiently scheduling these requests becomes an important component of LLM serving.

A common baseline is First-Come, First-Served (FCFS) scheduling, which processes requests in arrival order. However, FCFS can suffer from head-of-line blocking~\citep{kaffes2019shinjuku}: a long-running request can delay many shorter requests queued behind it, increasing average completion time. Size-based scheduling addresses this problem by prioritizing shorter requests. Policies such as Shortest Job First (SJF) and Shortest Remaining Processing Time (SRPT) are well studied in scheduling theory and, under standard assumptions, minimize average completion time.

Applying size-based scheduling to LLM inference presents a fundamental challenge: the size of a request depends largely on its output length, which is unknown when the request arrives. Recent LLM serving systems address this challenge by predicting request size from the input prompt and using these predictions to guide scheduling decisions~\citep{shahout2025don, fu2024efficient, qiu2024efficient, choi2025elis, zheng2026uncertainty, wen2026outlets}. For example, TRAIL predicts output length from the prompt's intermediate representations~\citep{shahout2025don}, while LTR~\cite{fu2024efficient} predicts the relative ordering of request lengths. TRAIL then uses its length estimates in an SRPT-style scheduler.

This reliance on predictions derived from user-controlled prompts introduces a new attack surface. An adversary can modify its prompt to reduce the predicted output length while preserving the original request, causing the scheduler to assign the request higher priority (\secref{sec:threat-model}). Unlike attacks that increase an LLM request's computational demand (e.g.,~\citet{zhang2025crabs}), our goal is to create a mismatch between the scheduler's predicted and actual request sizes: the scheduler predicts a short request and assigns it high priority, while the request still generates a substantially longer response.

In this paper, we introduce \attack{}, an attack on prediction-based LLM schedulers that exploits this vulnerability. \attack{} constructs an adversarial suffix that, when appended to a request, causes the scheduler's length predictor to underestimate its output length, thereby increasing the request's scheduling priority.

We study \trail{}~\cite{shahout2025don} as a case study because its scheduling pipeline exposes the trust boundary at the center of our attack: the user controls the request, TRAIL predicts its output length from the LLM's internal representation, and the scheduler uses this prediction to assign priority. Specifically, TRAIL uses a lightweight probe~\citep{belinkov2022probing} to map the request's internal representation to an output-length estimate. \attack{} targets this prediction, causing adversarial requests to receive higher scheduling priority without requiring the attacker to modify the serving system, scheduler, or other users' requests.

To construct the adversarial suffix, \attack{} directly optimizes the scheduler's predicted output length. It uses gradient-guided token optimization~\citep{zou2023universal} to iteratively modify the suffix, searching for tokens that reduce the predicted length. We then evaluate whether the resulting suffixes preserve the request's response length and quality, and whether they transfer across models.

We evaluate \attack{} across four LLMs and multiple request distributions, examining its effect on length predictions, scheduling behavior, and response utility. \attack{} consistently reduces predicted output lengths across models and tasks, demonstrating that adversarial suffixes can successfully manipulate the scheduling signal. This manipulation causes attacked requests to receive higher scheduling priority: in our end-to-end serving experiments, attacked requests move substantially earlier in the completion order and complete up to \(1.53\times\) faster on average. Importantly, the reduction in predicted length is substantially larger than the change in actual output length, demonstrating that \attack{} creates a mismatch between the scheduler's estimate and the request's realized size. Response quality varies across models and tasks, revealing a trade-off between scheduling advantage and response utility. We further show that adversarial suffixes can transfer across models and evaluate scheduler-side defenses that reduce the attack's scheduling advantage.

Looking forward, our work points to a broader attack surface in prediction-driven LLM inference systems. Beyond length-based scheduling, recent systems use model-derived predictions to guide optimization decisions, such as allocating inference-time compute or orchestrating reasoning branches~\citep{fu2024efficiently,jiang2025intra,wen2026outlets,brown2026predictive}. Understanding whether and how these prediction-driven optimizations can be manipulated through user-controlled inputs presents an important direction for future work.

\section{Related Work}

\subsection{LLM Serving and Prediction-Based Scheduling}

LLM serving systems improve efficiency through continuous batching and iteration-level scheduling~\citep{yu2022orca}, memory-efficient KV-cache management~\citep{kwon2023efficient}, throughput-latency optimization~\citep{agrawal2024taming}, prefill-decode disaggregation~\citep{zhong2024distserve}, dynamic load
balancing~\citep{sun2024llumnix}, phase-splitting
architectures~\citep{patel2024splitwise}, KV-cache-centric storage~\citep{qin2025mooncake}, and iteration-level preemption~\citep{wu2026fastserve}. These systems generally make scheduling decisions from observable run-time state, including queue occupancy, KV-cache pressure, and generated-token progress.
Recently, some line of work predicts request cost before or during generation and uses that prediction to approximate shortest-job-first or shortest-remaining-processing-time scheduling. Qiu
et al.~\citep{qiu2024efficient} use a proxy model to predict output length before generation, while Fu et al.~\citep{fu2024efficient} learn the relative
ordering of request lengths. TRAIL~\citep{shahout2025don} obtains an initial length estimate from the
target LLM's intermediate representations, refines the estimate as tokens are
generated, and uses it in an SRPT-style scheduler with limited preemption.
ELIS~\citep{choi2025elis} instead uses a separate encoder-based predictor and
periodically updates its remaining-length estimates during iterative
scheduling. We study TRAIL because it exposes the trust boundary central to our question: a user-controlled request determines an internal representation, a learned probe maps that representation to a length estimate, and the scheduler
directly uses that estimate to assign priority.

\subsection{Adversarial and Resource-Consumption Attacks on LLMs}

Gradient-guided discrete optimization over token sequences originates with
HotFlip~\citep{ebrahimi2018hotflip} and underlies transferable adversarial suffix attacks such as GCG~\citep{zou2023universal}, with subsequent work
improving attack stealth, universality, and transferability \citep{liu2024autodan,ben2026universal}. 
These attacks primarily target the model's generative behavior, for example by eliciting a harmful or policy-violating completion. We use a related discrete optimization procedure but define its loss solely on the scheduler-facing length predictor.
A second line of work studies resource-consumption and latency attacks. Sponge examples increase the energy or latency of neural-network inference
\citep{shumailov2021sponge}, while LLM-specific attacks induce unusually long,
repetitive, or non-terminating generations
\citep{gao2024denial,zhang2025crabs,li2025thinktrap,li2026loopllm}. These attacks increase the request's realized computational demand. By contrast, our attack does not try to increase the number of tokens generated. We make the scheduler underestimate how many tokens the model will generate, creating a mismatch between predicted and actual cost.
We then test whether this mismatch changes the request's scheduling position and its effect on co-scheduled benign traffic.

\subsection{Defenses and Robust Scheduling}

Defenses against adversarial suffixes have largely focused on content-safety failures, including randomized input perturbation and aggregation
\citep{robey2023smoothllm}, backtranslation-based
verification~\citep{wang2024defending}, robust prompt
optimization~\citep{zhou2024robust}, and representation-level jailbreak detection~\citep{zhang2025jbshield}. 
Their objective is to prevent unsafe completions, whereas our setting requires protecting the integrity of a scheduler-facing prediction.
Work on robust scheduling assumes that service-time predictions contain ordinary prediction errors~\citep{mitzenmacher2020scheduling,zheng2026uncertainty}; it does not consider users deliberately changing their inputs to mislead the
predictor. Outside LLM serving, Wu et al.~\citep{wu2022scalable} show that the
GNN-based cluster scheduler Decima is not always strategy-proof: a user can sometimes get a job scheduled earlier by reporting misleading job attributes. By contrast, this work study a similar problem in LLM serving, where the attacker modifies the request text to make a learned predictor underestimate the response length. We evaluate whether this gives the attacker higher scheduling priority and
whether scheduler-side defenses can prevent it.

\section{Methodology}

\subsection{System and Threat Model}
\label{sec:system-model}
\label{sec:threat-model}

\paragraph{Prediction-based scheduling.}
We study an LLM serving system that predicts the response length of each request and uses this prediction to determine scheduling priority. Following TRAIL~\citep{shahout2025don}, we define a \emph{request} as the complete user-controlled input submitted for one model response. For a request \(x\), the server extracts the hidden states from transformer layer \(\ell\) and averages them across the input tokens:
\(h_\ell(x) = \operatorname{MeanPool}\left(\operatorname{Layer}_\ell(x)\right).\)
A learned probe \(g_\theta\) maps \(h_\ell(x)\) to a distribution over \(K\) response-length bins. Let \(c_k\) denote the representative length of bin \(k\). The predicted response length is
\(\hat{n}(x) = \sum_{k=1}^{K} \operatorname{softmax}\left(g_\theta(h_\ell(x))\right)_k c_k.\)
The scheduler uses \(\hat{n}(x)\) as its initial estimate of the request's service time. Requests with smaller estimates receive higher priority. This design creates an attack surface: the user controls the request \(x\), and the prediction derived from that request determines its scheduling priority. A user may alter \(x\) to lower this prediction while the model continues to generate many tokens, allowing the request to receive higher priority.

\paragraph{Threat model.}
The attacker can append an \(m\)-token suffix
\(\delta \in \mathcal{V}^{m}\) to its own request before submission. During suffix construction, the attacker has white-box access to the target model and the length probe, including gradients through the model up to layer \(\ell\) and through \(g_\theta\). The model parameters, probe parameters, scheduling policy, and other users' requests remain fixed. At deployment, the attacker submits the modified request \(x \oplus \delta\). We optimize a separate suffix for each request.

We also consider cross-model transfer. There, the attacker constructs a suffix using a source model and its length probe, then submits the suffix with the corresponding request to a receiving model. The receiving model's probe and generated response provide the measurements of transferability.

\subsection{JIL Attack}
\label{sec:jil-attack}

JIL constructs the suffix by minimizing the predicted response length. Given a request \(x\) and an \(m\)-token suffix \(\delta \in \mathcal{V}^{m}\), the attack solves
\(\delta^\star = \arg\min_{\delta \in \mathcal{V}^{m}} \mathcal{L}_{\mathrm{sched}}(x,\delta), \qquad \mathcal{L}_{\mathrm{sched}}(x,\delta) = \hat{n}(x \oplus \delta),\)
where \(x \oplus \delta\) denotes the request with the suffix appended. Here, \(\mathcal{L}_{\mathrm{sched}}\) denotes the optimization objective, and \(\hat{n}\) denotes the predicted length used to evaluate each candidate suffix.

We optimize the suffix using Greedy Coordinate Gradient (GCG)~\citep{zou2023universal}. At each step, JIL computes the gradient of \(\mathcal{L}_{\mathrm{sched}}\) with respect to the token embeddings at each suffix position. It uses these gradients to select candidate replacement tokens, evaluates the resulting suffixes, and keeps the candidate with the lowest predicted length. Gradients pass through the model up to layer \(\ell\) and through the length probe, directing the optimization toward the scheduler-facing prediction.

\subsection{Evaluation Criteria}
\label{sec:evaluation-criteria}

We evaluate JIL at three levels. We first measure the change in predicted length, then examine service times and completion positions in the serving system, and finally evaluate generated response lengths and task utility. Together, these measurements connect the attack objective to scheduling behavior and response quality.

\paragraph{Prediction manipulation.}
For a clean request \(x\) and its attacked version \(x \oplus \delta\), we define the absolute prediction change as
\(\Delta_{\mathrm{pred}}(x) = \hat{n}(x \oplus \delta)-\hat{n}(x).\)
A negative value means that the suffix lowers the predicted length. Since clean predictions differ across requests and models, we also report the per-request relative reduction
\(R_{\mathrm{pred}}(x) = \frac{\hat{n}(x)-\hat{n}(x \oplus \delta)}{\hat{n}(x)}.\)
A larger \(R_{\mathrm{pred}}\) indicates a stronger reduction relative to the original prediction.

\paragraph{Service time.}
We evaluate scheduling using paired clean and attacked executions of each workload. Within workload \(w\), let \(G\) denote either the target-request group or the benign-request group. For request \(i\), let \(T_{w,i}^{\mathrm{first}}\) denote time to first token and \(T_{w,i}^{\mathrm{done}}\) denote completion latency, both measured from request submission.

For either time metric \(T\), we compute the group-mean time in each condition \(c \in \{\mathrm{clean},\mathrm{attacked}\}\):
\(\overline{T}_{w,G}^{\,c} = \frac{1}{|G|} \sum_{i \in G} T_{w,i}^{\,c}.\)
We then define the workload-level speedup factor as
\(S_{w,G}^{T} = \frac{\overline{T}_{w,G}^{\,\mathrm{clean}}}{\overline{T}_{w,G}^{\,\mathrm{attacked}}}.\)
A speedup above one indicates faster service under attack, a value of one indicates equal mean times, and a value below one indicates slower service. We compute completion-time and TTFT speedups separately for target and benign requests. We summarize these workload-level ratios across the paired workloads, with each workload contributing one observation to the corresponding boxplot.

\paragraph{Completion position.}
Service times describe how quickly requests receive their first token and finish. Completion positions describe their ordering within the workload. Let \(r_{w,i}\) denote the completion rank of request \(i\), with rank one assigned to the first request to finish, and let \(N_w\) denote the number of requests in workload \(w\). We define the normalized completion position as
\(p_{w,i}^{\,c} = \frac{r_{w,i}^{\,c}}{N_w}.\)
The paired position change, expressed in percentage points, is
\(\Delta p_{w,i} = 100\left(p_{w,i}^{\,\mathrm{attacked}} - p_{w,i}^{\,\mathrm{clean}}\right).\)
A negative value indicates movement toward the front of the completion order, and a positive value indicates movement toward the back. We average these changes within each request group and workload, then summarize the workload-level means. This normalization expresses position changes relative to the size of the workload.

\paragraph{Benign-request delays.}
We examine individual benign requests to measure how delays are distributed within each workload. For completion latency, we define the paired slowdown factor
\(D_{w,i} = \frac{T_{w,i}^{\mathrm{done,attacked}}}{T_{w,i}^{\mathrm{done,clean}}}.\)
A value above one indicates that the request finishes more slowly under attack. For each workload, we report the fraction of benign requests with \(D_{w,i}>1\) and the 95th percentile of their slowdown factors. We then average each measure across workloads. The slower-request fraction describes how widely delays occur, while the 95th percentile describes the upper tail of their relative magnitude. We use the same service-time, completion-position, and delay measures when evaluating early stopping and defenses.

\paragraph{Response utility.}
Let \(n_{\mathrm{actual}}(x)\) denote the number of tokens generated for request \(x\). We compare \(n_{\mathrm{actual}}(x)\) before and after the attack to measure changes in generated response length. We then measure response utility using a task-specific evaluation: judged response quality for open-ended generation and answer accuracy for multiple-choice tasks. We also report repetition, malformed-output rate, and length-cap rate to describe output characteristics. Together, these measurements show how scheduling gains relate to changes in response length and usefulness.

\section{Experiments}
\label{sec:experiments}

We evaluate JIL through six research questions that follow the criteria defined in \secref{sec:evaluation-criteria}.
\textbf{RQ1 (Prediction manipulation):} How much does JIL reduce predicted response lengths across models and
request distributions?
\textbf{RQ2 (Scheduling impact):}
Does JIL make attacked requests run earlier, and does this delay benign requests in the same queue?
\textbf{RQ3 (Response utility):} How does JIL affect actual output length and task utility across instruction-following, open-ended chat, and multiple-choice tasks?
\textbf{RQ4 (Attack-strength trade-off):} How does stopping JIL earlier change response utility and scheduling advantage?
\textbf{RQ5 (Cross-model transferability):} How much do suffixes constructed on one model change predictions on another model?
\textbf{RQ6 (Defenses):} How do prediction floors and prediction grouping change target service and benign delays?

\subsection{Experimental Setup}
\label{sec:experimental-setup}

\noindent\textbf{Models and probes.}
We evaluate Llama-3-8B-Instruct, Qwen2.5-7B-Instruct, Mistral-7B-Instruct-v0.3, and Llama-3.1-70B-Instruct. Each model has a separately trained length probe using the same architecture and training protocol. \appref{app:models} provides the probe architecture and training data.

\noindent\textbf{Datasets and generation.}
For each model, we evaluate 200 Alpaca requests, 200 UltraChat requests, 100 ARC-Challenge examples, and 100 HellaSwag examples. Open-ended generation uses greedy decoding with a 512-token limit; multiple-choice generation uses a 16-token limit. \appref{app:models} describes dataset construction and generation settings.

\noindent\textbf{Attack configuration.}
JIL optimizes a 20-token suffix over 30 GCG steps, evaluating 64 candidate suffixes per step. RQ4 evaluates earlier suffix selection at different prediction-reduction thresholds. \appref{app:earlystop} provides the initialization, optimization, and selection details.

\noindent\textbf{Response evaluation.}
For Alpaca, an LLM judge compares each generated answer with the matching dataset answer. For UltraChat, the judge assigns scores from 1 to 5 and compares clean and attacked answers in randomized order. ARC-Challenge and HellaSwag use generated-answer accuracy. \appref{app:response-metrics} defines these measures and the additional output diagnostics.

\noindent\textbf{Paired scheduling evaluation.}
We evaluate JIL in the TRAIL/vLLM serving system using Llama-3-8B-Instruct on one NVIDIA B200 GPU. Each of five workloads contains 20 benign and five target requests. Clean and attacked executions preserve request identities, roles, submission order, and arrival times; target requests receive JIL suffixes in the attacked condition. We evaluate each workload with 16 and 64 sequence slots. \appref{app:paired-protocol} provides the execution protocol.

\noindent\textbf{Transfer and defense configurations.}
We evaluate bidirectional suffix transfer between Llama-3-8B and Mistral-7B on 200 paired requests. We also evaluate prediction floors and prediction grouping on Llama-3-8B using the five paired workloads at both capacity settings. RQ5 and RQ6 present the configurations; \appref{app:transfer}, \tabref{tab:app-floors}, and \appref{app:quantization} provide the detailed protocols.

\noindent\textbf{Statistical reporting.}
Prediction and utility comparisons use request-level paired bootstrap 95\% confidence intervals. Scheduling results summarize five paired workloads using the metrics defined in \secref{sec:evaluation-criteria}, with 95\% \(t\)-intervals. For prediction grouping, we average three execution repeats within each workload before computing the summary statistics. \appref{app:paired-protocol} provides details.

\subsection{RQ1: Prediction Manipulation}
\label{sec:prediction-manipulation}

\begin{wrapfigure}[16]{r}{0.60\textwidth}
\vspace{-0.8\baselineskip}
\centering
\captionsetup{font=small,skip=2pt}
\begin{subfigure}[t]{0.50\linewidth}
    \centering
    \includegraphics[width=\linewidth]{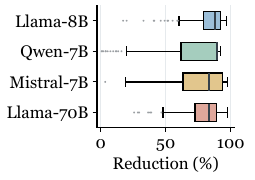}
    \caption{Alpaca.}
    \label{fig:prediction-alpaca}
\end{subfigure}
\hspace{-0.025\linewidth}
\begin{subfigure}[t]{0.50\linewidth}
    \centering
    \includegraphics[width=\linewidth]{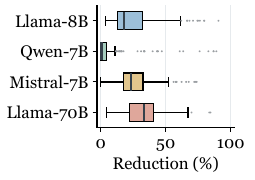}
    \caption{UltraChat.}
    \label{fig:prediction-ultrachat}
\end{subfigure}
\caption{Per-request relative reduction in predicted response length (\(n=200\) per model). Boxes show medians and quartiles; whiskers extend to the most extreme observations within 1.5 interquartile ranges, and points
show the remaining observations. JIL produces large reductions on Alpaca, while UltraChat shows stronger dependence on the model.}
\label{fig:prediction-distributions}
\vspace{-8mm}
\end{wrapfigure}

We first examine whether JIL consistently reduces the output of the length probe across models and request distributions. \figref{fig:prediction-distributions}
shows the request-level distributions for Alpaca and UltraChat. Complete means, medians, and reduction rates are reported in \appref{app:prediction}, \tabref{tab:app-relative}.

On Alpaca, JIL substantially reduces predicted response lengths across all four models. \figref{fig:prediction-distributions}(a) shows high median reductions, with the central half of requests also showing large decreases. The lower-reduction observations indicate variation across requests, while the distributions show that the effect extends across models, including Llama-3.1-70B.
On UltraChat, the distributions separate more clearly by model. \figref{fig:prediction-distributions}(b) shows positive median reductions for Llama-3-8B, Mistral-7B, and Llama-3.1-70B, with smaller reductions than on Alpaca. Qwen2.5-7B differs: its median lies close to zero, while some requests show much larger reductions. Its mean reduction consequently reflects an effect concentrated in a smaller subset of requests. Comparing the two panels shows that the same attack procedure produces different reduction distributions across both models and datasets.
The multiple-choice experiments extend this comparison to requests with short answers. JIL lowers the mean predicted length for every model on both ARC-Challenge and HellaSwag. Appendix \tabref{tab:app-mc} reports the clean and attacked predictions for each model and benchmark. These results extend the observed prediction manipulation from open-ended generation to multiple-choice tasks.

\begin{wrapfigure}[15]{r}{0.64\textwidth}
\vspace{-1\baselineskip}
\centering
\captionsetup{font=footnotesize,skip=1pt}
\captionsetup[subfigure]{font=scriptsize,skip=1pt}
\begin{subfigure}[t]{0.315\linewidth}
\centering
\includegraphics[width=\linewidth,height=0.60\linewidth]{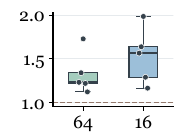}
\caption{Tar Completion speedup.}
\label{fig:paired-target-completion}
\end{subfigure}
\hspace{0.006\linewidth}
\begin{subfigure}[t]{0.315\linewidth}
\centering
\includegraphics[width=\linewidth,height=0.60\linewidth]{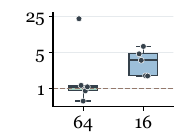}
\caption{Tar TTFT speedup.}
\label{fig:paired-target-ttft}
\end{subfigure}
\hspace{0.006\linewidth}
\begin{subfigure}[t]{0.315\linewidth}
\centering
\includegraphics[width=\linewidth,height=0.60\linewidth]{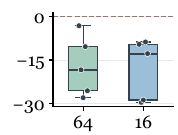}
\caption{Tar Position change.}
\label{fig:paired-target-position}
\end{subfigure}
\par\vspace{-0.2\baselineskip}
\begin{subfigure}[t]{0.315\linewidth}
\centering
\includegraphics[width=\linewidth,height=0.60\linewidth]{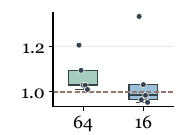}
\caption{Ben completion speedup.}
\label{fig:paired-benign-completion}
\end{subfigure}
\hspace{0.006\linewidth}
\begin{subfigure}[t]{0.315\linewidth}
\centering
\includegraphics[width=\linewidth,height=0.60\linewidth]{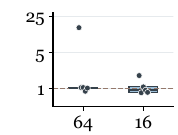}
\caption{Ben TTFT speedup.}
\label{fig:paired-benign-ttft}
\end{subfigure}
\hspace{0.006\linewidth}
\begin{subfigure}[t]{0.315\linewidth}
\centering
\includegraphics[width=\linewidth,height=0.60\linewidth]{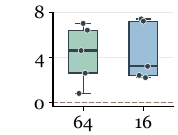}
\caption{Ben position change.}
\label{fig:paired-benign-position}
\end{subfigure}
\caption{Relative scheduling effects across five paired workloads. Tar refers to Target, Ben refers to Benign. Each point represents one workload; boxes show medians and quartiles, and whiskers extend to observations within 1.5 interquartile ranges.}
\label{fig:paired-scheduling}
\vspace{-4mm}
\end{wrapfigure}

\subsection{RQ2: Scheduling Impact}
\label{sec:scheduling-impact}

We next examine the resulting service times and completion order in the live TRAIL/vLLM serving system. Each clean run is paired with an attacked run that preserves request identities, roles, submission order, and planned arrival times. Both capacity settings contain 25 requests; the low-pressure setting provides 64 sequence slots, and the constrained setting provides 16 slots. \appref{app:paired-protocol} provides the complete paired protocol.

\figref{fig:paired-scheduling}(a) shows that target requests complete faster under attack in every paired workload under both capacity settings. The 16-slot setting has a larger mean completion-time speedup, with greater variation across workloads. The completion advantage persists across the two settings and becomes larger on average when fewer requests can run concurrently.
The first-token measurements show a different pattern. With 16 slots, target TTFT speedups exceed one in every workload, indicating earlier first-token delivery under attack. With 64 slots, most workload values lie near one, while a single large speedup raises the mean. \figref{fig:paired-scheduling}(b) shows why the workload distribution matters: the larger mean TTFT speedup at 64 slots is driven by one workload, whereas the 16-slot setting shows earlier first-token delivery across all five workloads.
The completion-position panels show how the targets' progress relates to the benign requests. Targets move forward in the completion order under both settings, while benign requests move backward on average within every workload (\figref{fig:paired-scheduling}(c,f)). Because the paired executions contain the same requests, their group-weighted position changes balance. Earlier target positions correspond to later benign positions.

The benign timing measurements show how these ordering changes translate into elapsed time. \figref{fig:paired-scheduling}(d,e) places most benign completion-time and TTFT speedups near one, with effects varying across workloads. At the request level, delays are more frequent and the upper-tail slowdown is larger with 16 slots. Within that setting, benign requests in the earlier half of the clean completion order become slower more frequently than those in the later half. \appref{app:scheduling} provides these request-level analyses. The workload averages thus accompany a distribution in which some benign requests experience increased completion times.
Completion times also reflect the amount of generated work and how requests are batched. Targets generate fewer tokens under attack in both capacity settings, so the measured speedups combine changes in scheduling predictions with changes in response length. Appendix Tables~\ref{tab:app-controls}--\ref{tab:app-mechanism} examine the contributions of request text, predicted length, generated length, and scheduling policy.

\subsection{RQ3: Output Length and Response Utility}
\label{sec:response-utility}

\begin{table}[H]
\centering
\small
\captionsetup{font=small,skip=3pt}
\renewcommand{\arraystretch}{0.98}
\vspace{-2mm}
\caption{Task utility before and after JIL. Each entry reports clean $\rightarrow$ attacked. Alpaca reports the win rate against the matching dataset answer, UltraChat reports the mean judge score on a 1--5 scale, and MC reports accuracy averaged over ARC-Challenge and
HellaSwag.}
\label{tab:cross-task-main}
\begin{tabular*}{\columnwidth}{
    @{\extracolsep{\fill}}lccc@{}}
\toprule
Model
& Alpaca (\%)
& UltraChat
& MC (\%) \\
\midrule
Llama-3-8B
& $84.0 \rightarrow 23.0$
& $3.42 \rightarrow 3.09$
& $72.0 \rightarrow 68.5$ \\
Qwen2.5-7B
& $93.5 \rightarrow 63.0$
& $3.49 \rightarrow 3.23$
& $82.5 \rightarrow 83.0$ \\
Mistral-7B
& $89.0 \rightarrow 74.0$
& $3.35 \rightarrow 3.17$
& $71.5 \rightarrow 62.5$ \\
Llama-3.1-70B
& $80.5 \rightarrow 55.5$
& $3.23 \rightarrow 2.43$
& $88.5 \rightarrow 86.5$ \\
\bottomrule
\end{tabular*}
\vspace{-0.4\baselineskip}
\end{table}

\begin{wrapfigure}[18]{r}{0.58\textwidth}
\vspace{-1.2\baselineskip}
\centering
\captionsetup{font=small,skip=2pt}
\begin{subfigure}[t]{0.50\linewidth}
\centering
\includegraphics[width=\linewidth]
{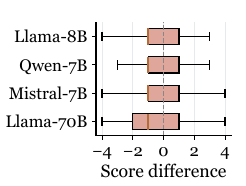}
\caption{Per-request score differences.}
\label{fig:utility-score-differences}
\end{subfigure}
\hspace{-0.025\linewidth}
\begin{subfigure}[t]{0.50\linewidth}
\centering
\includegraphics[width=\linewidth]
{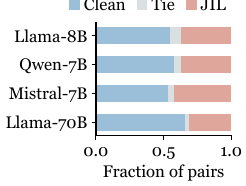}
\caption{Paired judge preferences.}
\label{fig:utility-preferences}
\end{subfigure}
\caption{UltraChat utility over 200 paired requests per model. Panel (a) shows the distribution of attacked-minus-clean judge scores. Boxes span the interquartile range, and orange lines mark medians. The median equals $-1$ for all four models and coincides with the lower quartile for Llama-8B, Qwen-7B, and Mistral-7B. Panel (b) shows the fractions of comparisons assigned to the clean response,
a tie, or the attacked response.}
\label{fig:utility-ultrachat-distributions}
\vspace{-0.8\baselineskip}
\end{wrapfigure}

We next examine how JIL affects generated response length and utility. On Alpaca and UltraChat, attacked responses remain several hundred tokens long on average, and their relative length decreases are smaller than the corresponding prediction reductions. Appendix \tabref{tab:app-output} reports the output lengths and diagnostics. We then evaluate the usefulness of these responses using the task-specific measures in \tabref{tab:cross-task-main}.
On Alpaca, reference win rates decrease for all four models. Llama-3-8B has the largest decrease despite producing nearly the same mean output length in both conditions. This combination shows that responses can retain their length while losing utility. The paired confidence intervals for all four win-rate changes lie below zero; Appendix \tabref{tab:app-utility-ci} provides the intervals.

\figref{fig:utility-ultrachat-distributions} examines utility across individual UltraChat requests. In panel (a), the median attacked-minus-clean score is negative for every model, while the boxes span both score decreases and increases. The attack's effect on quality varies across requests, with a lower score at the median. Panel (b) shows the corresponding preference results: the judge favors clean responses more frequently than attacked responses for every model. Llama-3.1-70B has both the largest mean score decrease and the strongest preference for clean responses. Mistral-7B has the smallest mean decrease, with a confidence interval spanning zero.
Multiple-choice accuracy changes less for three models, while Mistral-7B shows a larger decrease on both ARC-Challenge and HellaSwag. Appendix \tabref{tab:app-mc} reports the benchmark-specific results. Together, these measurements show that utility loss depends on the model and task, even as attacked responses retain substantial output length.

\subsection{RQ4: Attack-Strength Trade-off?}
\label{sec:early-stopping}

\begin{wrapfigure}[16]{r}{0.54\textwidth}
\vspace{-1.2\baselineskip}
\centering
\captionsetup{font=small,skip=2pt}
\begin{subfigure}[t]{0.48\linewidth}
\centering
\includegraphics[width=\linewidth]{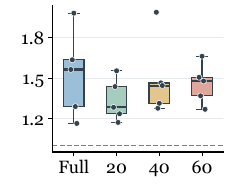}
\caption{Target requests.}
\label{fig:early-stop-target}
\end{subfigure}
\hfill
\begin{subfigure}[t]{0.48\linewidth}
\centering
\includegraphics[width=\linewidth]{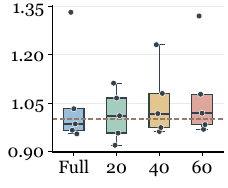}
\caption{Benign requests.}
\label{fig:early-stop-benign}
\end{subfigure}
\caption{Completion-time speedups for the full 30-step attack and three early-stop variants at 16 slots. Each point is the clean group-mean time divided by the attacked group-mean time for one paired workload. Boxes show medians and quartiles; whiskers extend to observations within 1.5 interquartile ranges.}
\label{fig:early-stop-scheduling}
\vspace{-0.4\baselineskip}
\end{wrapfigure}

We next examine whether selecting suffixes at smaller prediction reductions improves response utility while retaining scheduling gains. For each of the 200 Llama-3-8B Alpaca requests, we select the first suffix that lowers the clean prediction by 20, 40, or 60 tokens, using the final-step suffix when the threshold remains unreached. We compare these variants with the full 30-step attack on the same five paired workloads under the 16-slot setting. \appref{app:earlystop} provides the selection procedure.
Larger thresholds produce larger mean relative prediction reductions, with all three variants below the full attack. \figref{fig:early-stop-scheduling}(a) shows that each variant retains a target completion speedup above one in every workload. The 20-token variant has the smallest mean speedup, while the 40- and 60-token variants remain closer to the full attack. Targets also move forward in the completion order under all three variants. Earlier suffix selection reduces prediction manipulation while preserving a completion advantage.

For benign requests, \figref{fig:early-stop-scheduling}(b) shows overlapping completion-speedup distributions, with workload values on both sides of one. The request-level measurements show a smaller mean upper-tail slowdown for each early-stop variant than for the full attack. The smallest tail slowdown occurs at the intermediate threshold, while the smallest target speedup occurs at the lowest threshold. The effects on target completion and benign delays follow different patterns as the threshold changes.
Surprisingly, response utility decreases further under earlier suffix selection. All three variants have lower Alpaca reference win rates than the full attack, and the lowest threshold combines the smallest prediction reduction with the lowest win rate. Appendix \tabref{tab:app-stopping} reports the prediction and utility measurements. These variants retain scheduling gains while yielding lower response utility, showing that smaller prediction reductions can still accompany substantial quality loss.

\subsection{RQ5: Cross-Model Transferability}
\label{sec:transferability}

\begin{wraptable}{r}{0.58\textwidth}
\centering
\small
\setlength{\tabcolsep}{3pt}
\renewcommand{\arraystretch}{1.08}
\vspace{-4mm}
\caption{Cross-model transfer on 200 paired requests. Arrows indicate source $\rightarrow$ receiving model; Llama and Mistral denote Llama-3-8B and Mistral-7B. Prediction reductions are computed per request using the receiving model's probe. Output lengths are means in tokens.}
\vspace{-3mm}
\label{tab:transfer-main}
\begin{tabular*}{\linewidth}{@{\extracolsep{\fill}}lrr@{}}
\toprule
Measure
& Llama $\rightarrow$ Mistral
& Mistral $\rightarrow$ Llama \\
\midrule
Mean reduction (\%)   & 58.8  & 25.2  \\
Median reduction (\%) & 60.7  & 21.4  \\
Clean output tokens  & 396.3 & 426.2 \\
Transfer output tokens & 345.2 & 377.6 \\
\bottomrule
\end{tabular*}
\end{wraptable}

We next test whether suffixes optimized for one model also reduce another model's length predictions. We evaluate transfer in both directions between Llama-3-8B and Mistral-7B using 200 paired requests. Each receiving model processes the clean request and the request with the transferred suffix through its own probe and generator.
\tabref{tab:transfer-main} shows prediction reductions in both directions. Both the mean and median reduction are larger when Mistral-7B receives Llama-3-8B suffixes. In the same evaluation, suffixes optimized directly on Mistral-7B achieve a larger mean reduction of \(74.8\%\). Prediction manipulation transfers between the models, with its strength depending on the transfer direction.
Both receiving models generate shorter responses with transferred suffixes, while their mean output lengths remain several hundred tokens. The relative decreases in mean output length are smaller than the mean per-request prediction reductions. Overall, transfer reproduces the pattern observed in the direct attacks: predicted lengths decrease more strongly than generated lengths. \appref{app:transfer} provides the full protocol and measurements.

\subsection{RQ6: Defenses}
\label{sec:defenses}

We next examine whether changing the predictions supplied to the scheduler reduces target speedups and benign delays. We evaluate two rules: a prediction floor raises estimates below a minimum, and prediction grouping rounds estimates up to fixed-width token intervals. Both rules apply to target and benign requests in the clean and attacked conditions. \tabref{tab:defenses-main} summarizes their effects under the 16-slot setting, with each rule compared against its own batch's original-prediction baseline.

\begin{table}[htbp]
\centering\small
\renewcommand{\arraystretch}{0.98}
\vspace{-1mm}
\caption{Defenses at 16 sequence slots. Target speedup uses clean/attacked completion time; position gains are normalized by 25 requests and expressed in percentage points. Benign tail is the mean within-workload 95th percentile of attacked/clean request latency. Slower is the percentage of benign requests with increased latency. Each block uses its own batch's original-prediction baseline.}
\vspace{-3mm}
\label{tab:defenses-main}
\begin{tabular*}{\columnwidth}{@{\extracolsep{\fill}}lrrrr@{}}
\toprule
Setting & Target speedup & Position gain & Benign tail & Slower (\%) \\
\midrule
Floor baseline & $1.43\times$ & 14.24 & $1.35\times$ & 37.0 \\
Floor 80 & $1.36\times$ & 12.96 & $1.26\times$ & 27.0 \\
Floor 120 & $1.20\times$ & 10.88 & $1.22\times$ & 59.0 \\
\midrule
Grouping baseline & $1.42\times$ & 15.25 & $1.43\times$ & 55.3 \\
Interval 32 & $1.42\times$ & 15.68 & $1.38\times$ & 50.7 \\
Interval 64 & $1.35\times$ & 12.96 & $1.21\times$ & 40.3 \\
Interval 128 & $1.25\times$ & 8.80 & $1.13\times$ & 44.0 \\
\bottomrule
\end{tabular*}
\vspace{-0.5\baselineskip}
\end{table}

Higher prediction floors reduce target completion speedups and position gains. They also reduce the upper-tail slowdown of benign requests. The fraction of benign requests experiencing delays follows a different pattern: it decreases with the lower floor and increases with the higher floor. The higher floor combines a smaller tail slowdown with delays affecting more benign requests. These measurements show how the magnitude and frequency of benign delays respond differently to the floor setting.
Coarser prediction grouping also reduces target completion speedups and position gains. Both coarser intervals lower benign tail slowdown and the fraction of slower benign requests relative to the baseline. The widest interval produces the smallest target advantage and benign tail slowdown, while the intermediate interval yields the smallest fraction of slower benign requests. At 64 slots, target completion speedups remain similar across grouping settings. The larger changes at 16 slots show that grouping has a stronger effect when fewer requests can run concurrently. \tabref{tab:app-floors} and \appref{app:quantization} provide the full results.

The evaluated defenses act on the predictions used for scheduling. Further defenses could address the probe itself through adversarial training on modified requests paired with measured response lengths. A consistency check could compare predictions across perturbed versions of a request and route large discrepancies to a fallback policy, adding predictor evaluations. At the scheduler level, incorporating waiting time or reserving a service share for queued requests could support benign progress as priorities change. These approaches offer complementary ways to address prediction sensitivity and request fairness.

\section{Conclusion}
\label{sec:conclusion}

JIL optimizes adversarial suffixes to lower length predictions and gain scheduling priority. Across four LLMs, open-ended responses retain substantial length as predictions decrease. Paired serving experiments show faster target completion and increased benign tail delays under constrained capacity. Utility losses vary across models and tasks; early stopping retains scheduling gains while further lowering Alpaca win rates. Suffixes also transfer between Llama-3-8B and Mistral-7B, reducing the receiving model's predictions. Prediction floors and coarse grouping reduce target advantages, with coarse grouping also reducing benign tail slowdown and the fraction of delayed benign requests. As learned predictions guide access to shared computation, their sensitivity to user-controlled inputs becomes a scheduling fairness issue.

\section*{AI Use Statement}

AI-assisted tools were used for grammar checking and presentation-quality review, including the identification of sentence fragments, dangling words, discontinuities, and formatting inconsistencies.
We also used coding agents to assist with code development. We reviewed the generated code and validated the experimental pipeline, among others, through syntax checks, small-scale smoke tests, and inspection of execution logs and saved outputs. We checked request matching, sample counts,
and consistency between recorded measurements and reported results. 
The authors reviewed and take full responsibility for the manuscript.

\section*{Ethics Statement}

This work studies how user-controlled prompts affect resource allocation in LLM serving systems. The serving experiments use researcher-configured TRAIL/vLLM instances and requests drawn from established benchmarks; target and benign requests are experimental roles within these workloads. The attack has dual-use implications: manipulating length predictions could give an adversary preferential service and increase other users' waiting times in shared deployments. We evaluate prediction floors and coarse prediction grouping as scheduler-side mitigations, measuring their effects on both target service and benign delays. These measurements support the design of serving policies that account for adversarial inputs and fairness among users. Follow-up security testing should use authorized deployments and controlled workloads.

\section*{Reproducibility Statement}

\secref{sec:jil-attack} specifies the suffix optimization, and \secref{sec:experimental-setup} describes the models and evaluation settings. \appref{app:protocol} details request construction, response measurements, paired workloads, and statistical aggregation. Apps.~\ref{app:prediction} and~\ref{app:trajectory} describe the transfer, generated-length probe, and prefix evaluations; Apps.~\ref{app:scheduling}--\ref{app:earlystop} provide the scheduling controls, utility measurements, attack variants, and defense settings. \appref{app:records} describes the saved request-level records and source, checkpoint, and input hashes used to trace the reported results. 
We also release implementation details at:
\url{https://anonymous.4open.science/r/jil-C2F8}.

\bibliography{references}

\clearpage
\appendix
\raggedbottom
\setlength{\textfloatsep}{10pt plus 2pt minus 2pt}
\setlength{\floatsep}{8pt plus 2pt minus 2pt}
\setlength{\intextsep}{8pt plus 2pt minus 2pt}
\section{Experimental Protocol and Measurements}
\label{app:protocol}

The supplementary experiments follow the sequence of measurements in the main paper. We first specify the request construction and evaluation protocol, then connect prediction reductions to generated lengths and cross-model transfer. Prefix evaluations track the prediction as a response grows. Scheduling controls separate request text, predicted length, generated work, and policy. Task-level utility measurements lead into comparisons of earlier stopping, a next-token objective, and scheduler-side prediction controls.

\subsection{Models, Probes, and Evaluation Requests}
\label{app:models}
We evaluate Llama-3-8B-Instruct, Qwen2.5-7B-Instruct,
Mistral-7B-Instruct-v0.3, and Llama-3.1-70B-Instruct. Each model has a separate probe with a 512-unit hidden layer and ten output bins. The probe receives a mean-pooled representation at layer 11. Its input dimension matches the model's hidden dimension. Training uses 2,000 Alpaca reference responses, and checkpoint selection uses 500 validation responses. Training response-length deciles define the bins; the mean length within each bin defines its representative value.
Each model is evaluated on 200 Alpaca requests, 200 UltraChat requests, 100 ARC-Challenge examples, and 100 HellaSwag examples. Alpaca sampling uses a fixed seed and a minimum reference-response length of 60 tokens.
UltraChat inputs use the first user turn of each selected conversation. Each multiple-choice example receives its own optimized suffix. Open-ended generation uses greedy decoding with a 512-token limit. Multiple-choice generation uses a 16-token limit followed by answer parsing.

\subsection{Prediction and Response Measurements}
\label{app:response-metrics}
For request $i$, let $\widehat L_i^c$ and $\widehat L_i^a$ denote the clean and attacked predictions. We compute the relative reduction for each pair, then summarize the resulting request-level distribution:
\begin{equation}
r_i=\frac{\widehat L_i^c-\widehat L_i^a}{\widehat L_i^c},\qquad
\overline r=\frac{1}{n}\sum_{i=1}^{n}r_i.
\end{equation}
The fraction reaching a threshold $q$ is
$n^{-1}\sum_i\mathbf{1}[r_i\geq q]$. The 20\%, 40\%, and 60\% thresholds describe this relative reduction. Early stopping uses the absolute-token thresholds defined in \appref{app:earlystop}.
Response evaluation retains the metric associated with each task. Alpaca compares each generated answer with the corresponding dataset answer. UltraChat compares clean and attacked answers to the same request. ARC-Challenge and HellaSwag score the parsed answer against the correct
option. \tabref{tab:app-metrics} defines these measures and the output diagnostics used alongside them.

\begin{table}[!htbp]
\centering\small
\caption{Response measures and their aggregation. Each response receives equal weight within its model--dataset condition.}
\label{tab:app-metrics}
\begin{tabular}{@{}p{.25\linewidth}p{.70\linewidth}@{}}
\toprule
Measure & Definition \\
\midrule
Alpaca win rate & Fraction of comparisons won by the generated answer against the matching dataset answer; ties remain in the denominator. \\
UltraChat score & Mean judge score, with endpoints 1 (unusable) and 5 (highly useful). \\
UltraChat preference & Fractions assigned clean win, attacked win, and tie in direct paired comparisons. \\
MC accuracy & Fraction of parsed generated answers matching the correct option. \\
Output length & Number of generated model-token IDs after the input. \\
Length-cap rate & Fraction satisfying the generation batch's recorded cap indicator. \\
Repeated 3-grams & Mean per-answer fraction of overlapping text 3-grams that repeat an earlier occurrence. \\
Malformed rate & Fraction flagged by an empty answer or a character fraction above 15\% with Unicode code point greater than 127. \\
\bottomrule
\end{tabular}
\end{table}

The UltraChat judge requests a preference and two integer scores in one evaluation. Answer order is randomized, and the returned answer labels are mapped to their experimental conditions. The Mistral, Qwen, and 70B UltraChat evaluations use \texttt{openai/gpt-5-mini} through OpenRouter. The recovered Mistral and 70B Alpaca evaluation manifests record \texttt{gpt-5-mini} and seed 42.
Saved judge records retain the model, provider, answer order, scores, preference, and error field.

For repetition, we lowercase each answer and tokenize it into word runs and individual punctuation characters. An answer with $m\geq3$ text units has $m-2$ overlapping 3-grams. If $U$ of them are distinct, its repetition
rate is $1-U/(m-2)$. Answers with fewer than three units receive zero. The malformed flag records an output-format property, while the task metrics measure answer quality. UltraChat exports compute the character fraction after trimming surrounding whitespace; the recovered Mistral
and 70B Alpaca exports use the original answer string and a trimmed emptiness check. The Mistral Alpaca cap flag marks lengths of at least 510 tokens. The 70B Alpaca and new small-model UltraChat generators mark exactly 512 tokens.

\subsection{Workloads and Statistical Calculation}
\label{app:paired-protocol}
Scheduling uses Llama-3-8B-Instruct on one NVIDIA B200 GPU. Five workloads each contain 20 benign and five target requests. Each workload is run under clean and attacked conditions with matched request identities, roles, submission order, and planned arrival times. The runner spaces planned arrivals by 10\,ms. The two capacity settings use 16 and 64 sequence slots for the same 25 requests.
The runner applies the model's chat template and submits token IDs through \texttt{AsyncLLMEngine}. It supplies the initial predicted length through \texttt{SamplingParams.remain\_length}. The scheduler uses chunked prefill and its LRPSPRPT policy. A capacity guard checks the active sequence budget before admitting another running group. Request records retain the input, prediction sent to the engine, submission timestamp, first-token timestamp, completion timestamp, output length, and completion rank.

TTFT and completion latency are measured from actual submission. For time metric $T$ and request group $G$ in workload $w$, the speedup is $S^T_{w,G}=\overline T^c_{w,G}/\overline T^a_{w,G}$. Values above one indicate faster service under attack. The normalized position change is $\Delta p_{w,i}=100(r^a_{w,i}-r^c_{w,i})/25$, expressed in percentage points. Negative changes indicate earlier completion positions. Group means give each request equal weight within its workload.

For individual benign requests, the completion slowdown is $D_{w,i}=T^a_{w,i}/T^c_{w,i}$. We calculate its 95th percentile and the fraction with $D_{w,i}>1$ within each workload, then average the five workload values. Speedup describes group-mean service; the slowdown distribution describes the magnitude and frequency of individual delays. These definitions are shared by the main results, early stopping, mechanism controls, and defenses.

Scheduling summaries average five workload observations and use 95\% $t$-intervals with four degrees of freedom. Repeated executions first produce a ratio, position change, tail percentile, and slower-request fraction for each workload and repeat. We average repeats within the same workload before computing the five-workload summary. Prediction and utility intervals resample matched request IDs; utility intervals use 10,000 paired bootstrap samples.

\section{Prediction Reduction and Generated Length}
\label{app:prediction}

\subsection{Relative Reduction Across Requests}
The model--dataset comparison shows how frequently a suffix produces a large reduction. \tabref{tab:app-relative} reports the mean, median, and threshold rates over 200 paired requests per cell. Alpaca reductions are large across all four models. UltraChat separates the models more
strongly: Qwen has a median reduction of 0.6\%, while the other medians range from 18.2\% to 32.9\%.
\begin{table}[!htbp]
\centering\small
\setlength{\tabcolsep}{4pt}
\renewcommand{\arraystretch}{1.10}
\caption{Per-request relative prediction reductions, with 200 pairs per model and dataset. All entries are percentages.}
\label{tab:app-relative}
\begin{tabular}{@{}llrrrrr@{}}\toprule
Dataset & Model & Mean & Median & $\geq20\%$ & $\geq40\%$ & $\geq60\%$ \\
\midrule
Alpaca & Llama-8B & 83.4 & 88.3 & 99.0 & 98.5 & 94.0 \\
Alpaca & Llama-70B & 79.3 & 83.2 & 100.0 & 97.0 & 89.5 \\
Alpaca & Mistral-7B & 76.6 & 83.5 & 99.0 & 95.5 & 78.5 \\
Alpaca & Qwen2.5-7B & 72.6 & 89.2 & 94.0 & 87.0 & 76.0 \\
UltraChat & Llama-8B & 25.3 & 18.2 & 47.5 & 18.0 & 6.5 \\
UltraChat & Llama-70B & 32.7 & 32.9 & 78.0 & 27.0 & 3.0 \\
UltraChat & Mistral-7B & 25.4 & 23.2 & 65.5 & 11.0 & 3.0 \\
UltraChat & Qwen2.5-7B & 6.9 & 0.6 & 9.0 & 5.5 & 3.5 \\
\bottomrule\end{tabular}
\end{table}

\subsection{Suffix Transfer Between Llama and Mistral}
\label{app:transfer}
The model-specific experiments optimize each suffix against the probe that evaluates it. We next apply the saved suffix text to a second model. Both transfer directions use the same 200 Alpaca requests. Llama-to-Mistral transfer uses the Llama suffixes from the supplementary $\lambda=0$ run. Mistral-to-Llama transfer uses suffixes optimized against the Mistral probe. The receiving model tokenizes the concatenated prompt and suffix, evaluates its own probe, and generates a response. Source optimization uses 20-token suffixes and 30 search steps; the receiving tokenizer determines the transferred text's token count.

\tabref{tab:app-transfer} reports the receiving model's prediction and generated length. Llama suffixes reduce Mistral predictions by 58.8\% on average, with a median reduction of 60.7\%. Mistral suffixes reduce Llama predictions by 25.2\% on average, with a median of 21.4\%. The transferred suffixes lower predictions in both directions, with a larger average effect when Mistral receives the Llama suffixes.

Mistral's own optimized suffixes produce a mean relative reduction of 74.8\%, compared with 58.8\% for the transferred Llama suffixes. Their mean generated lengths are 362.83 and 345.16 tokens, respectively. This comparison places the transferred prediction change alongside optimization on the receiving model.
\begin{table}[!htbp]
\centering\small
\setlength{\tabcolsep}{4pt}
\renewcommand{\arraystretch}{1.10}
\caption{Bidirectional suffix transfer on 200 paired requests. Prediction and length columns report receiving-model means in tokens; reduction is the mean per-request percentage.}
\label{tab:app-transfer}
\begin{tabular}{@{}lrrrrr@{}}\toprule
Direction & Pred. $c$ & Pred. $a$ & Reduction & Length $c$ & Length $a$ \\
\midrule
Llama to Mistral & 87.05 & 34.80 & 58.8\% & 396.33 & 345.16 \\
Mistral to Llama & 104.64 & 75.49 & 25.2\% & 426.24 & 377.60 \\
\bottomrule\end{tabular}
\end{table}

\subsection{Clean Predictions and Generated Responses}
The relative reduction describes the change in the probe output. To compare that output to generated responses, we pair each clean prediction with its observed generated length. \tabref{tab:app-calibration} reports error, signed bias, rank correlation, and the fraction reaching
512 tokens. Bias is prediction minus generated length. Across the eight model--dataset cells, mean bias ranges from $-347.2$ to $-277.5$ tokens on UltraChat and from $-350.1$ to $-309.1$ on Alpaca.
The probes are trained on reference-response lengths, while these measurements use the evaluated model's generated responses under the 512-token limit. The rank correlations describe how the clean predictions order the observed lengths. They range from 0.153 to 0.477 on Alpaca and
from 0.320 to 0.718 on UltraChat.

\begin{table}[!htbp]
\centering\small
\setlength{\tabcolsep}{4pt}
\renewcommand{\arraystretch}{1.10}
\caption{Clean predictions compared with generated token counts under the 512-token limit. Each row contains 200 requests. Bias is predicted minus generated length; the final column gives the percentage with 512 output tokens.}
\label{tab:app-calibration}
\begin{tabular}{@{}llrrrr@{}}\toprule
Dataset & Model & MAE & Bias & Spearman & At 512 (\%) \\
\midrule
Alpaca & Llama-8B & 350.46 & -350.08 & 0.153 & 70.0 \\
Alpaca & Qwen2.5-7B & 339.65 & -338.49 & 0.338 & 49.0 \\
Alpaca & Mistral-7B & 309.70 & -309.13 & 0.477 & 39.5 \\
Alpaca & Llama-70B & 328.79 & -328.41 & 0.279 & 49.0 \\
UltraChat & Llama-8B & 282.51 & -277.50 & 0.718 & 61.5 \\
UltraChat & Qwen2.5-7B & 347.80 & -347.20 & 0.320 & 65.5 \\
UltraChat & Mistral-7B & 326.77 & -325.81 & 0.665 & 63.0 \\
UltraChat & Llama-70B & 345.32 & -342.73 & 0.546 & 73.5 \\
\bottomrule\end{tabular}
\end{table}

\FloatBarrier
\subsection{Training on Model-Generated Response Lengths}
\label{app:generated-probe}
\begin{wraptable}{r}{0.52\textwidth}
\centering\small
\setlength{\tabcolsep}{4pt}
\renewcommand{\arraystretch}{1.10}
\caption{Llama-8B predictions and generated lengths on the 200 attack requests, using the probe trained on generated response lengths. All entries are token counts.}
\label{tab:app-generated-probe}
\begin{tabular}{@{}lrr@{}}\toprule
Measurement & Clean & Attacked \\
\midrule
Mean prediction & 415.65 & 47.83 \\
Mean generated length & 426.24 & 346.01 \\
\bottomrule\end{tabular}
\end{wraptable}

The preceding comparison pairs reference-trained predictions with model-generated responses. We next train a Llama-8B probe using lengths generated by Llama-8B itself. The model answers 3,000 Alpaca requests with greedy decoding and a 512-token cap. The split assigns 2,000 requests to training, 500 to validation, and 500 to testing. Request identities and prompt text are disjoint from the 200-request attack pool.

The probe uses the same layer-11 feature extraction and a 512-unit hidden layer with ten output bins. Training-length deciles define the bin boundaries. Each occupied bin uses its mean training length as its representative value; empty bins use the training median. AdamW trains the probe for 80 epochs with a learning rate of $10^{-3}$, weight decay of $10^{-4}$, and batches of 128. The checkpoint with the lowest validation MAE is evaluated on the test split.

The selected probe achieves a test MAE of 48.76 tokens, compared with 174.32 tokens for the constant training-median predictor. Its RMSE is 75.63 tokens and its Spearman correlation is 0.911. Bin accuracy is 42.6\%: the highest-probability bin matches the bin containing the observed generated length for 213 of the 500 test requests. Of these test responses, 19.6\% reach the 512-token cap. Together, the error, correlation, and bin accuracy describe numerical prediction error, request ordering, and length-bin classification against the generated-length target.

We then optimize a 20-token suffix for each of the 200 attack requests using 30 search steps. \tabref{tab:app-generated-probe} connects the resulting prediction changes with generated response lengths. The mean prediction falls from 415.65 to 47.83 tokens, with a mean per-request relative reduction of 86.3\% and a median of 90.4\%. Mean generated length changes from 426.24 to 346.01 tokens. The attack also changes the predictions of the probe trained on model-generated lengths.

\section{Prediction Changes During Generation}
\label{app:trajectory}

\subsection{Prefix Evaluation}
We evaluate the frozen Llama-8B probe on 200 clean--attacked response pairs at generated-token positions 0, 2, 8, 16, 32, 64, 128, 256, 384, and 512.
At each position, the probe receives the request and the corresponding generated prefix. We evaluate raw-input and chat-formatted sequences separately. The clean-minus-attacked prediction gap is
\begin{equation}
g_i(t)=\widehat L_i^c(t)-\widehat L_i^a(t).
\end{equation}
A positive gap places the attacked prediction below the clean prediction. The probe weights remain fixed throughout prefix evaluation.
The number of paired observations changes when either answer ends. All 200 pairs reach position 16; 197 reach position 64, 187 reach position 128, 141 reach position 256, 83 reach position 384, and 38 reach position
512. For comparisons across positions, we select a horizon $H$ and retain the pairs whose two responses both reach $H$. Every point on a horizon-specific curve uses that same group of requests.

\subsection{Evolution Within a Fixed Request Group}
\begin{figure}[!htbp]
\centering
\begin{subfigure}[t]{.48\linewidth}\centering
\includegraphics[width=\linewidth]{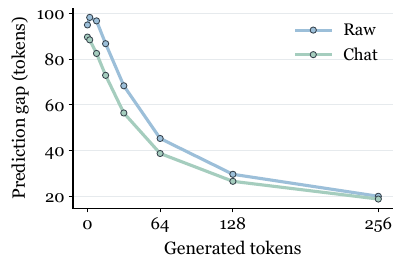}
\caption{Mean gap through generation.}
\end{subfigure}\hfill
\begin{subfigure}[t]{.48\linewidth}\centering
\includegraphics[width=\linewidth]{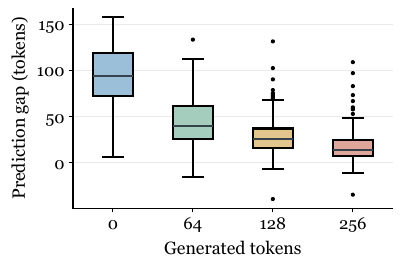}
\caption{Raw-input request distributions.}
\end{subfigure}
\caption{Frozen-probe prediction gaps for the fixed group of 141 pairs whose two responses reach 256 tokens. Positive values indicate a lower attacked prediction. Panel (a) follows raw and chat-formatted inputs; panel (b) shows medians, quartiles, 1.5-IQR whiskers, and outlying values
for raw inputs at selected positions.}
\label{fig:app-trajectory}
\end{figure}

\figref{fig:app-trajectory} follows the 141 pairs reaching 256 generated tokens. For raw inputs, the mean gap is 94.98 tokens at position 0, 45.25 at position 64, 29.56 at position 128, and 19.92 at position 256. For chat-formatted inputs, the corresponding means are 89.67, 38.67,
26.48, and 18.71 tokens. Both trajectories show a smaller gap as the generated prefix grows. The boxplots show how that change is distributed across the same 141 request pairs.

\section{Scheduling Outcomes and Mechanism Controls}
\label{app:scheduling}

\subsection{Completion Under the Paired Workloads}
\tabref{tab:app-absolute} reports the primary paired workloads using the same relative metrics as the main paper. Target completion speedups exceed one on average at both capacities, and the constrained setting gives the larger mean speedup. Target position changes are negative while benign position changes are positive. The timing and ordering measurements connect the target's faster completion to its movement within the queue.

Benign group-mean completion speedups are close to one. The request-level tail and slower-request fraction separate the two capacities more clearly: delays affect a larger fraction of benign requests at 16 slots and reach a larger relative magnitude in the upper tail. This contrast motivates controls that vary the scheduler input and generated work separately.

\begin{table}[!htbp]
\centering\small
\setlength{\tabcolsep}{4pt}
\renewcommand{\arraystretch}{1.10}
\caption{Relative service in the primary paired workloads. Completion speedup includes a 95\% workload-level interval. TTFT is a speedup factor; position is attacked minus clean in percentage points. Tail is the mean within-workload 95th-percentile slowdown; slower is a percentage.}
\label{tab:app-absolute}
\begin{tabular*}{\linewidth}{@{\extracolsep{\fill}}rlrrrrr@{}}\toprule
Slots & Role & Completion [95\% CI] & TTFT & Position & Tail & Slower \\
\midrule
64 & Target & 1.33 [1.03, 1.62] & 5.22 & -17.12 & 0.97 & 4.0 \\
64 & Benign & 1.07 [0.97, 1.17] & 3.82 & +4.28 & 0.97 & 6.0 \\
16 & Target & 1.53 [1.13, 1.93] & 3.66 & -17.92 & 0.97 & 8.0 \\
16 & Benign & 1.05 [0.86, 1.25] & 1.10 & +4.48 & 1.45 & 43.0 \\
\bottomrule\end{tabular*}
\end{table}

\subsection{Text, Prediction, Output Length, and Scheduling Policy}
The natural-output experiment changes the target's suffix and its scheduler-facing prediction together. We separate those inputs through the configurations in \tabref{tab:app-controls}. The signal control keeps the clean request text and changes the target prediction. The text control appends the suffix and retains the clean prediction. A fixed-output configuration generates 512 tokens per request. The FCFS configuration uses arrival-time ranking in the same serving engine.

\begin{table}[!htbp]
\centering\small
\caption{Controlled scheduling configurations. Entries describe the target in the attacked condition; its clean counterpart uses the clean text and prediction. All configurations retain 20 benign and five target requests per workload.}
\label{tab:app-controls}
\begin{tabular}{@{}llll@{}}\toprule
Configuration & Target text & Target prediction & Output \\
\midrule
Natural attack & With suffix & Attacked & Natural ending \\
Signal control & Clean & Attacked & Natural ending \\
Text control & With suffix & Clean & Natural ending \\
Fixed output & With suffix & Attacked & 512 tokens \\
Fixed-output signal & Clean & Attacked & 512 tokens \\
\bottomrule\end{tabular}
\end{table}

The control batch contains 16 configurations and 4,000 request executions, including the floor settings in \appref{app:earlystop}. \tabref{tab:app-control-times} reports the natural-output, fixed-output, signal, and text comparisons. Natural-output targets have a completion advantage under both prediction-based scheduling and FCFS. With fixed output, completion speedups move closer to one. The signal and text controls then locate the service changes associated with each input to the execution: the prediction controls queue priority, and the suffix can also change the generated response.

\begin{table}[!htbp]
\centering\small
\setlength{\tabcolsep}{4pt}
\renewcommand{\arraystretch}{1.10}
\caption{Relative service across controlled configurations. Times are reported as clean/attacked group-mean ratios. Target completion includes a 95\% interval across five workloads; benign is completion speedup.}
\label{tab:app-control-times}
\begin{tabular*}{\linewidth}{@{\extracolsep{\fill}}rlrrr@{}}\toprule
Slots & Configuration & Target TTFT & Target completion [95\% CI] & Benign \\
\midrule
16 & Natural, prediction & 3.25 & 1.43 [0.99, 1.88] & 1.00 \\
16 & Natural, FCFS & 1.34 & 1.22 [0.99, 1.44] & 1.01 \\
16 & Fixed, prediction & 6.05 & 1.05 [0.92, 1.18] & 1.00 \\
16 & Fixed, FCFS & 1.03 & 1.00 [0.98, 1.03] & 1.00 \\
16 & Signal control & 1.97 & 1.10 [1.01, 1.19] & 0.99 \\
16 & Text control & 0.97 & 1.18 [0.93, 1.43] & 1.04 \\
64 & Natural, prediction & 3.96 & 1.31 [1.00, 1.62] & 1.05 \\
64 & Natural, FCFS & 1.89 & 1.27 [1.02, 1.51] & 1.01 \\
64 & Fixed, prediction & 1.16 & 1.00 [0.97, 1.03] & 1.00 \\
64 & Fixed, FCFS & 1.73 & 1.02 [0.94, 1.09] & 1.02 \\
64 & Signal control & 1.34 & 0.99 [0.97, 1.01] & 1.00 \\
64 & Text control & 4.52 & 1.32 [0.97, 1.67] & 1.04 \\
\bottomrule\end{tabular*}
\end{table}

\subsection{Fixed-Output Signal Comparison with FCFS}
We next combine the clean-text signal control with fixed output length. Both conditions use the clean request text and generate 512 tokens. The target prediction changes from its clean value to its attacked value. Five workloads run under two policies and two capacities, with three repeats per configuration, giving twelve configurations and 3,000 request executions.

For each workload and policy, we first average the three execution-level speedups. The relative policy effect is $C^T_{w,G}=\overline S^{T,\mathrm{pred}}_{w,G}/\overline S^{T,\mathrm{FCFS}}_{w,G}$. A value above one indicates a larger attack-induced speedup under prediction-based scheduling. The position contrast is $C^p_{w,G}=\overline{\Delta p}^{\mathrm{pred}}_{w,G}-\overline{\Delta p}^{\mathrm{FCFS}}_{w,G}$. Negative values indicate greater forward movement under the prediction policy. \tabref{tab:app-mechanism} summarizes these workload-level comparisons.

\begin{figure}[!htbp]\centering
\begin{subfigure}[t]{.48\linewidth}\centering
\includegraphics[width=\linewidth]{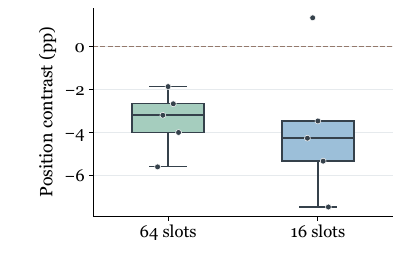}
\caption{Normalized position contrast.}
\end{subfigure}\hfill
\begin{subfigure}[t]{.48\linewidth}\centering
\includegraphics[width=\linewidth]{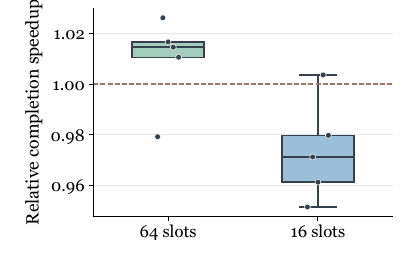}
\caption{Relative completion speedup.}
\end{subfigure}
\caption{Policy comparisons for targets with clean request text and 512 generated tokens. Each point summarizes three repeats of one workload. Panel (a) subtracts the FCFS normalized position change from the prediction-policy change; negative values indicate greater forward movement. Panel (b) divides the prediction-policy completion speedup by the FCFS speedup; values above one indicate a larger speedup. Boxes show medians and quartiles, with 1.5-IQR whiskers.}

\label{fig:app-mechanism}
\end{figure}

The target position contrast is negative at both capacities. Four of five workloads have a negative contrast at 16 slots, and all five do at 64 slots. \figref{fig:app-mechanism} pairs this ordering result with the relative completion-speedup distribution. The position changes are accompanied by completion-speedup contrasts closer to one, connecting changes in queue order to their timing effect when generated work is held fixed.

\begin{table}[!htbp]
\centering\small
\setlength{\tabcolsep}{4pt}
\renewcommand{\arraystretch}{1.10}
\caption{Policy comparisons with clean request text and 512 generated tokens. Relative speedup divides the prediction-policy speedup by the FCFS speedup after averaging three repeats within each workload. Position contrast subtracts the FCFS normalized position change in percentage points.}
\label{tab:app-mechanism}
\begin{tabular*}{\linewidth}{@{\extracolsep{\fill}}rlrrr@{}}\toprule
Slots & Role & Relative TTFT & Relative completion [95\% CI] & Position contrast \\
\midrule
16 & Target & 1.20 & 0.97 [0.95, 1.00] & -3.84 \\
16 & Benign & 0.97 & 0.98 [0.96, 1.00] & +0.96 \\
64 & Target & 1.28 & 1.01 [0.99, 1.03] & -3.47 \\
64 & Benign & 1.07 & 1.01 [0.99, 1.03] & +0.87 \\
\bottomrule\end{tabular*}
\end{table}

\subsection{Changes Within the Benign Request Group}
Every workload assigns completion ranks 1 through 25. Their sum is the same in both conditions, giving $5\overline{\Delta p}_{\mathrm{target}}+20\overline{\Delta p}_{\mathrm{benign}}=0$. The fixed-output policy contrasts retain this balance: target contrasts of $-3.84$ and $-3.47$ percentage points accompany benign contrasts of $+0.96$ and $+0.87$ points at 16 and 64 slots.

We further divide the benign requests in each primary workload into equal halves by clean completion rank. At 16 slots, 52\% of the earlier half and 34\% of the later half finish more slowly under attack. At 64 slots, the corresponding fractions are 8\% and 4\%. The clean-order grouping identifies which benign requests more frequently experience increased completion time as targets move forward.

\section{Task Utility and Output Diagnostics}
\label{app:utility}

\subsection{Open-Ended Responses}
\label{app:openended-details}
The scheduling results describe service order and elapsed time. We now examine the answers produced by the attacked requests. \tabref{tab:app-output} reports output length and malformed-output flags for Alpaca and UltraChat. The attacked Alpaca means range from 270.76 to 448.14 tokens, and the attacked UltraChat means range from 388.95 to 408.11 tokens after rounding to two decimals.
\begin{table}[!htbp]
\centering\small
\setlength{\tabcolsep}{4pt}
\renewcommand{\arraystretch}{1.10}
\caption{Generated output measurements for 200 pairs per model and dataset. Lengths are mean token counts; malformed is the attacked-condition flag rate.}
\label{tab:app-output}
\begin{tabular}{@{}llrrr@{}}\toprule
Dataset & Model & Clean length & Attacked length & Malformed \\
\midrule
Alpaca & Llama-8B & 454.72 & 448.14 & 5.5\% \\
Alpaca & Qwen2.5-7B & 407.95 & 270.76 & 2.0\% \\
Alpaca & Mistral-7B & 395.56 & 367.68 & 0.5\% \\
Alpaca & Llama-70B & 409.77 & 361.52 & 4.5\% \\
UltraChat & Llama-8B & 395.66 & 388.95 & 0.0\% \\
UltraChat & Qwen2.5-7B & 417.18 & 403.87 & 11.0\% \\
UltraChat & Mistral-7B & 416.36 & 408.11 & 0.0\% \\
UltraChat & Llama-70B & 436.11 & 398.22 & 6.5\% \\
\bottomrule\end{tabular}
\end{table}

Task utility adds an assessment of answer content.
\tabref{tab:app-utility-ci} reports paired utility changes with bootstrap intervals. Alpaca win-rate changes range from $-61.0$ to $-15.0$ percentage points. UltraChat score changes range from $-0.80$ to $-0.18$. The largest decreases occur for Llama-8B on Alpaca and Llama-70B on UltraChat. Each row retains its task's metric and unit.
\begin{table}[!htbp]
\centering\small
\setlength{\tabcolsep}{4pt}
\renewcommand{\arraystretch}{1.10}
\caption{Paired utility changes with 95\% bootstrap intervals from 10,000 resamples of 200 request pairs. Changes are attacked minus clean.}
\label{tab:app-utility-ci}
\begin{tabular}{@{}llrr@{}}\toprule
Measure & Model & Change & 95\% interval \\
\midrule
Alpaca win (pp) & Llama-8B & -61.00 & [-68.00, -54.00] \\
UltraChat score & Llama-8B & -0.33 & [-0.54, -0.13] \\
Alpaca win (pp) & Qwen2.5-7B & -30.50 & [-37.50, -23.50] \\
UltraChat score & Qwen2.5-7B & -0.26 & [-0.45, -0.07] \\
Alpaca win (pp) & Mistral-7B & -15.00 & [-22.00, -8.50] \\
UltraChat score & Mistral-7B & -0.18 & [-0.37, 0.01] \\
Alpaca win (pp) & Llama-70B & -25.00 & [-33.50, -16.50] \\
UltraChat score & Llama-70B & -0.80 & [-1.06, -0.54] \\
\bottomrule\end{tabular}
\end{table}

\subsection{Multiple-Choice Answers}
\label{app:mc-details}
Multiple-choice evaluation makes the answer comparison explicit through the correct option. \tabref{tab:app-mc} gives the two benchmarks separately, with 100 clean--attacked pairs in each model--benchmark cell. The mean prediction falls in all eight cells. Accuracy changes range
from a one-percentage-point increase for Qwen on ARC-Challenge to a ten-point decrease for Mistral on the same benchmark. This breakdown shows the benchmark contributions to the main table's pooled accuracy.
\begin{table}[!htbp]
\centering\small
\setlength{\tabcolsep}{4pt}
\renewcommand{\arraystretch}{1.10}
\caption{Benchmark-level results, with 100 paired examples per row. Prediction columns are means in tokens; accuracy columns are percentages.}
\label{tab:app-mc}
\begin{tabular}{@{}llrrrr@{}}\toprule
Model & Benchmark & Pred. $c$ & Pred. $a$ & Acc. $c$ & Acc. $a$ \\
\midrule
Llama-8B & ARC-Challenge & 10.87 & 6.83 & 84\% & 79\% \\
Llama-8B & HellaSwag & 10.49 & 7.53 & 60\% & 58\% \\
Qwen2.5-7B & ARC-Challenge & 54.78 & 28.16 & 89\% & 90\% \\
Qwen2.5-7B & HellaSwag & 61.66 & 33.36 & 76\% & 76\% \\
Mistral-7B & ARC-Challenge & 14.92 & 6.25 & 76\% & 66\% \\
Mistral-7B & HellaSwag & 8.65 & 4.72 & 67\% & 59\% \\
Llama-70B & ARC-Challenge & 12.53 & 4.66 & 95\% & 94\% \\
Llama-70B & HellaSwag & 7.08 & 4.02 & 82\% & 79\% \\
\bottomrule\end{tabular}
\end{table}

\subsection{Concise-Response Prompt Baselines}
\label{app:prompt-baselines}
We also evaluate two concise-response instructions on the same 200 Llama-8B Alpaca requests. Both instructions achieve 181 wins against the matching dataset references, corresponding to a 90.5\% win rate. Their mean output lengths are 344.99 and 314.72 tokens, compared with
454.72 for clean requests and 448.14 for JIL. \tabref{tab:app-prompts} reports these response-level comparisons.
\begin{table}[!htbp]
\centering\small
\setlength{\tabcolsep}{4pt}
\renewcommand{\arraystretch}{1.10}
\caption{Llama-8B prompt baselines on the same 200 Alpaca requests. Wins compare each answer with the matching dataset reference.}
\label{tab:app-prompts}
\begin{tabular}{@{}lrrr@{}}\toprule
Condition & Wins / 200 & Win rate & Mean tokens \\
\midrule
Clean & 168 & 84.0\% & 454.72 \\
JIL & 46 & 23.0\% & 448.14 \\
Concise, high-quality & 181 & 90.5\% & 344.99 \\
Brief, complete, useful & 181 & 90.5\% & 314.72 \\
\bottomrule\end{tabular}
\end{table}

\section{Attack Objectives and Scheduler Prediction Controls}
\label{app:earlystop}
\label{app:threshold-details}

\subsection{Stopping at a Prediction-Reduction Threshold}
The utility results motivate a comparison across optimization endpoints. The Llama-8B implementation initializes the suffix by tokenizing a string of 20 exclamation marks, padding with exclamation-mark token IDs to 20 tokens, and truncating to that length. Each of 30 search steps uses embedding gradients to rank 64 replacements per suffix position. A seeded shuffle selects up to 64 one-position candidates, and the candidate with the lowest predicted length becomes the next suffix. Saved steps retain the decoded suffix and its prediction.

For each request, we select the first saved suffix whose prediction is 20, 40, or 60 tokens below its clean prediction. A trajectory that ends before reaching its threshold contributes its final suffix. The resulting mean relative reductions increase with the threshold. \tabref{tab:app-stopping} links these endpoints to reference win rates and relative scheduling measurements on the same five workloads at 16 slots.

All three endpoints retain target completion speedups above one, with the intermediate threshold producing the largest mean speedup among the early-stop variants. Their benign upper-tail slowdowns are smaller than the full attack's value. Reference win rates are also lower than the full attack's rate. The service and utility measurements describe how selecting an earlier suffix changes both outcomes.

\begin{table}[!htbp]
\centering\small
\setlength{\tabcolsep}{4pt}
\renewcommand{\arraystretch}{1.10}
\caption{Early-stopping outcomes. Prediction reduction and win rate use 200 requests. Scheduling uses five 25-request workloads at 16 slots. Speedup is clean/attacked target completion time, position gain is in percentage points, and benign tail is a request-level slowdown factor.}
\label{tab:app-stopping}
\begin{tabular*}{\linewidth}{@{\extracolsep{\fill}}lrrrrr@{}}\toprule
Endpoint & Reduction & Win rate & Speedup & Position gain & Benign tail \\
\midrule
20 tokens & 28.8\% & 10.5\% & 1.34 & 13.28 & 1.42 \\
40 tokens & 46.3\% & 14.0\% & 1.50 & 17.28 & 1.30 \\
60 tokens & 62.2\% & 13.5\% & 1.46 & 16.00 & 1.36 \\
30 steps & 83.4\% & 23.0\% & 1.53 & 17.92 & 1.45 \\
\bottomrule\end{tabular*}
\end{table}

\subsection{Adding a Next-Token Distribution Term}
\label{app:kl-objective}
Early stopping selects an endpoint using the predicted length. We also evaluate an objective that combines predicted length with a next-token distribution term. For raw prompt $x$ and suffix $s$, the objective is
\begin{equation}
\mathcal{J}_{\lambda}(s;x)=\widehat L(x+s)
 +\lambda\sum_{v\in\mathcal V}p(v\mid x)
 \log\frac{p(v\mid x)}{p(v\mid x+s)}.
\end{equation}
The sum covers the full model vocabulary at the final position of each raw input. Each search runs for 30 steps with a 20-token suffix, 64 gradient-ranked replacements per position, and up to 64 sampled replacement candidates per step. Candidate scoring decodes the suffix and tokenizes the complete prompt--suffix text. Response generation subsequently applies the chat template and a 512-token cap.

Each of the four configurations contains 200 requests. The $\lambda=0$ and $\lambda=0.05$ runs yield mean relative prediction reductions of 85.8\% and 85.8\%, respectively. Mean output lengths are 331.50 and 329.89 tokens, compared with 426.24 tokens for clean responses. \tabref{tab:app-kl-predictions} reports prediction changes and generated lengths across the four configurations.

The $\lambda=0.5$ and $\lambda=5$ runs each yield a mean prediction of 153.21 tokens, compared with 104.64 tokens for the clean inputs. Their mean per-request reduction is $-74.9\%$, indicating an increase. Mean generated length is 415.91 tokens in both configurations.
\begin{table}[!htbp]
\centering\small
\setlength{\tabcolsep}{4pt}
\renewcommand{\arraystretch}{1.10}
\caption{Supplementary objective runs on 200 Llama-8B requests per configuration. Prediction and output columns are means in tokens; reduction is the mean per-request percentage. Clean predictions average 104.64 tokens and clean outputs average 426.24 tokens.}
\label{tab:app-kl-predictions}
\begin{tabular}{@{}r r r r@{}}\toprule
$\lambda$ & Prediction & Reduction & Output \\
\midrule
0 & 14.52 & 85.8\% & 331.50 \\
0.05 & 14.59 & 85.8\% & 329.89 \\
0.5 & 153.21 & -74.9\% & 415.91 \\
5 & 153.21 & -74.9\% & 415.91 \\
\bottomrule\end{tabular}
\end{table}

We submit each saved suffix and prediction pool to the paired scheduler at 16 and 64 slots. Each configuration uses five workloads of 20 benign and five target requests, giving 2,000 request executions across eight configurations. \tabref{tab:app-kl-scheduling} reports completion speedups, normalized positions, and benign tail slowdowns. The two lower-weight pools accelerate targets and move them forward on average at both capacities. The two higher-weight pools have mean target completion speedups below one and positive mean position changes. These measurements connect the saved prediction endpoints to their service outcomes.

\begin{table}[!htbp]
\centering\small
\setlength{\tabcolsep}{4pt}
\renewcommand{\arraystretch}{1.10}
\caption{Relative scheduling outcomes. Completion is clean/attacked group-mean time; target position is attacked minus clean in percentage points. Benign tail is the 95th-percentile request slowdown. Entries average five workloads.}
\label{tab:app-kl-scheduling}
\begin{tabular*}{\linewidth}{@{\extracolsep{\fill}}rrrrrr@{}}\toprule
Slots & $\lambda$ & Target completion & Target position & Benign completion & Benign tail \\
\midrule
16 & 0 & 1.53 & -13.60 & 1.04 & 1.21 \\
16 & 0.05 & 1.42 & -12.64 & 0.99 & 1.36 \\
16 & 0.5 & 0.90 & +3.84 & 1.00 & 1.44 \\
16 & 5 & 0.98 & +1.28 & 1.02 & 1.35 \\
64 & 0 & 1.45 & -22.56 & 1.04 & 1.00 \\
64 & 0.05 & 1.28 & -17.44 & 1.02 & 1.02 \\
64 & 0.5 & 0.99 & +2.88 & 1.01 & 1.06 \\
64 & 5 & 0.99 & +3.36 & 1.02 & 1.06 \\
\bottomrule\end{tabular*}
\end{table}

\tabref{tab:app-kl-service} follows the same endpoints through first-token service and the fraction of individual requests that finish more slowly. TTFT speedups describe the beginning of service, while completion-delay fractions describe how frequently the total response time increases. Reading these results alongside the completion speedups shows how group averages and individual outcomes vary across the saved pools.

\begin{table}[!htbp]
\centering\small
\setlength{\tabcolsep}{4pt}
\renewcommand{\arraystretch}{1.10}
\caption{First-token speedups and individual delays. TTFT is clean/attacked group-mean time; slower is the percentage of requests with attacked/clean completion latency above one. Entries average five workloads.}
\label{tab:app-kl-service}
\begin{tabular*}{\linewidth}{@{\extracolsep{\fill}}rrrrrr@{}}\toprule
Slots & $\lambda$ & Target TTFT & Benign TTFT & Target slower (\%) & Benign slower (\%) \\
\midrule
16 & 0 & 2.48 & 1.06 & 4.0 & 27.0 \\
16 & 0.05 & 2.32 & 0.98 & 16.0 & 53.0 \\
16 & 0.5 & 0.63 & 1.14 & 64.0 & 55.0 \\
16 & 5 & 0.92 & 1.16 & 68.0 & 32.0 \\
64 & 0 & 1.25 & 1.31 & 4.0 & 17.0 \\
64 & 0.05 & 2.08 & 1.50 & 12.0 & 27.0 \\
64 & 0.5 & 1.16 & 1.28 & 52.0 & 36.0 \\
64 & 5 & 1.88 & 1.81 & 52.0 & 37.0 \\
\bottomrule\end{tabular*}
\end{table}

\subsection{Applying a Minimum Scheduler Prediction}
Early stopping changes the suffix selected by the attacker. A prediction floor changes the value supplied to the scheduler. For floor $F$, the scheduler receives
\begin{equation}
\widehat L^{\mathrm{sent}}=
\max\{F,\operatorname{round}(\widehat L)\}.
\end{equation}
We apply floors of 80 and 120 tokens to request predictions under both clean and attacked conditions and retain the paired workload design.

\tabref{tab:app-floors} compares each floor with the natural-output configuration from the same batch. At 16 slots, increasing the floor lowers target completion speedup and normalized position gain. The benign upper-tail slowdown also decreases. At 64 slots, the target completion speedups remain closer across floor settings. The difference between capacities connects the floor's effect to the number of requests competing for active sequence slots.

\begin{table}[!htbp]
\centering\small
\setlength{\tabcolsep}{4pt}
\renewcommand{\arraystretch}{1.10}
\caption{Relative service under prediction floors. Target TTFT and completion use clean/attacked time; position gain is in percentage points. Benign tail is the mean within-workload 95th-percentile slowdown. Intervals summarize five workload speedups.}
\label{tab:app-floors}
\begin{tabular*}{\linewidth}{@{\extracolsep{\fill}}rlrrrr@{}}\toprule
Slots & Floor & Target TTFT & Target completion [95\% CI] & Position gain & Benign tail \\
\midrule
16 & Original & 3.25 & 1.43 [0.99, 1.88] & 14.24 & 1.35 \\
16 & 80 tokens & 5.65 & 1.36 [1.20, 1.53] & 12.96 & 1.26 \\
16 & 120 tokens & 1.11 & 1.20 [0.97, 1.44] & 10.88 & 1.22 \\
64 & Original & 3.96 & 1.31 [1.00, 1.62] & 17.92 & 1.00 \\
64 & 80 tokens & 1.35 & 1.25 [0.99, 1.51] & 16.96 & 1.02 \\
64 & 120 tokens & 1.32 & 1.27 [0.98, 1.57] & 16.32 & 0.99 \\
\bottomrule\end{tabular*}
\end{table}

We then compare benign service across floor settings within the same clean or attacked condition. For each workload, the cross-setting speedup divides the original-prediction mean time by the corresponding floor-setting mean time. \tabref{tab:app-floor-overhead} reports the five-workload means and intervals. This comparison measures the service cost or benefit of applying the defense, alongside the attack-induced speedups measured within each floor setting.

\begin{table}[!htbp]
\centering\small
\setlength{\tabcolsep}{4pt}
\renewcommand{\arraystretch}{1.10}
\caption{Benign completion speedup after applying a floor. Each workload compares its original-prediction mean time with its floor-setting mean time in the same condition. Values above one indicate faster completion with the floor.}
\label{tab:app-floor-overhead}
\begin{tabular*}{\linewidth}{@{\extracolsep{\fill}}rrlr@{}}\toprule
Slots & Floor & Condition & Original/floor time [95\% CI] \\
\midrule
16 & 80 & Clean & 1.02 [0.99, 1.06] \\
16 & 80 & Attacked & 1.07 [1.01, 1.13] \\
16 & 120 & Clean & 1.03 [1.02, 1.04] \\
16 & 120 & Attacked & 1.02 [0.95, 1.08] \\
64 & 80 & Clean & 0.98 [0.94, 1.03] \\
64 & 80 & Attacked & 0.96 [0.94, 0.98] \\
64 & 120 & Clean & 0.98 [0.93, 1.03] \\
64 & 120 & Attacked & 0.97 [0.95, 0.98] \\
\bottomrule\end{tabular*}
\end{table}

\subsection{Grouping Predictions into Token Intervals}
\label{app:quantization}
A floor changes predictions below a minimum value. Grouping predictions into intervals also changes the spacing between the remaining values. For interval width $q>0$ and the integer prediction $\ell$ supplied by the runner, we use
\begin{equation}
Q_q(\ell)=\max\{q,\;q\lceil\ell/q\rceil\}.
\end{equation}
The setting $q=0$ retains the runner's original value. Widths of 32, 64, and 128 tokens are applied to every benign and target request in both conditions. Each of the two capacities uses five paired workloads and three execution repeats per width. The resulting 24 configurations contain 6,000 request executions. We average the three repeats within each workload and then average the five workloads.

\tabref{tab:app-quantization} reports relative completion times, normalized position changes, and benign tail slowdowns. At 16 slots, the wider intervals reduce the mean target completion speedup and forward position change. The widest interval gives the smallest target speedup and benign tail slowdown. At 64 slots, target completion speedups remain similar across widths. The larger change under constrained capacity links prediction grouping to competition for active slots.

\begin{table}[!htbp]
\centering\small
\setlength{\tabcolsep}{4pt}
\renewcommand{\arraystretch}{1.10}
\caption{Relative scheduling outcomes. Completion is clean/attacked group-mean time; target position is attacked minus clean in percentage points. Benign tail is the 95th-percentile request slowdown. Entries average three repeats within each workload and then five workloads.}
\label{tab:app-quantization}
\begin{tabular*}{\linewidth}{@{\extracolsep{\fill}}rrrrrr@{}}\toprule
Slots & $q$ & Target completion & Target position & Benign completion & Benign tail \\
\midrule
16 & 0 & 1.42 & -15.25 & 1.01 & 1.43 \\
16 & 32 & 1.42 & -15.68 & 0.98 & 1.38 \\
16 & 64 & 1.35 & -12.96 & 1.01 & 1.21 \\
16 & 128 & 1.25 & -8.80 & 1.02 & 1.13 \\
64 & 0 & 1.27 & -18.24 & 1.02 & 1.04 \\
64 & 32 & 1.27 & -17.60 & 1.01 & 1.06 \\
64 & 64 & 1.27 & -18.45 & 1.00 & 1.05 \\
64 & 128 & 1.25 & -17.60 & 1.01 & 1.06 \\
\bottomrule\end{tabular*}
\end{table}

\tabref{tab:app-quantization-service} adds first-token speedups and slower-request fractions using the same repeat and workload weighting. At 16 slots, the 64-token interval has the smallest fraction of slower benign requests, while the 128-token interval has the smallest benign tail slowdown. The two delay measures locate different effects of interval width: how widely increased latency occurs and how large its upper tail becomes.

\begin{table}[!htbp]
\centering\small
\setlength{\tabcolsep}{4pt}
\renewcommand{\arraystretch}{1.10}
\caption{First-token speedups and individual delays. TTFT is clean/attacked group-mean time; slower is the percentage of requests with attacked/clean completion latency above one. Entries average three repeats within each workload and then five workloads.}
\label{tab:app-quantization-service}
\begin{tabular*}{\linewidth}{@{\extracolsep{\fill}}rrrrrr@{}}\toprule
Slots & $q$ & Target TTFT & Benign TTFT & Target slower (\%) & Benign slower (\%) \\
\midrule
16 & 0 & 3.49 & 1.01 & 12.0 & 55.3 \\
16 & 32 & 3.84 & 0.96 & 17.3 & 50.7 \\
16 & 64 & 2.05 & 1.05 & 13.3 & 40.3 \\
16 & 128 & 1.28 & 1.08 & 16.0 & 44.0 \\
64 & 0 & 1.35 & 1.21 & 13.3 & 31.0 \\
64 & 32 & 1.65 & 1.52 & 12.0 & 37.3 \\
64 & 64 & 1.13 & 0.99 & 14.7 & 47.7 \\
64 & 128 & 1.22 & 1.08 & 16.0 & 36.7 \\
\bottomrule\end{tabular*}
\end{table}

\section{Result Records and Reproduction}
\label{app:records}
\label{app:provenance}
Each attack record stores its request ID, input, suffix, clean prediction, and attacked prediction. Saved optimization histories link these endpoints to the intermediate suffixes. Generation records add the answer, generated-token count, and output diagnostics. Judge records preserve the evaluated answer pair and returned judgment.
Scheduling records add request roles, arrival plans, timestamps, and completion ranks. These identifiers link the measurements used in each paired comparison.

The Llama-8B prediction replay reproduces all 400 saved clean and attacked predictions for the 200-request pool. The scheduling supplement contains 4,000 request executions across 16 configurations. The repeated fixed-output signal comparison contains 3,000 executions across twelve
configurations. The prefix evaluation contains 200 response pairs with position-specific coverage. Source hashes, checkpoint hashes, input hashes, manifests, and completion records accompany these results.

The additional batch contains 41 completion records: four objective runs, eight paired scheduling configurations for their saved endpoints, two transfer runs, generated-label collection, probe training, probe attack evaluation, and 24 prediction-grouping configurations. Per-request JSON records link each suffix to its clean and attacked predictions and generated answers. The objective runs, transfer evaluations, and probe attack each contain 200 request records. The label collection contains 3,000 requests, and the selected generated-length probe has 500 held-out test predictions.

\bibliographystyle{iclr2027_conference}

\end{document}